\documentclass[10pt,twocolumn,letterpaper]{article}

\usepackage{wacv}              

\usepackage{times}
\usepackage{epsfig}
\usepackage{graphicx}
\usepackage{amsmath}
\usepackage{amssymb}
\usepackage{booktabs}
\usepackage[linesnumbered,ruled,vlined]{algorithm2e}
\usepackage{amsmath}
\usepackage{algpseudocode}
\usepackage{gensymb}
\usepackage{diagbox}
\usepackage{xcolor}
\usepackage{multirow}
\usepackage{mathrsfs}
\usepackage{dsfont}
\usepackage{xcolor,colortbl}
\usepackage{float}
\usepackage[accsupp]{axessibility} 

\newcommand{\encoder}{\mathcal{E}}
\newcommand{\MLPA}{\mathcal{P}_A}
\newcommand{\MLPG}{\mathcal{P}_G}
\newcommand{\im}{\mathbf{x}}
\newcommand{\desc}{\mathbf{z}}

\usepackage[pagebackref,breaklinks,colorlinks]{hyperref}

\usepackage[capitalize]{cleveref}
\crefname{section}{Sec.}{Secs.}
\Crefname{section}{Section}{Sections}
\Crefname{table}{Table}{Tables}
\crefname{table}{Tab.}{Tabs.}

\begin{document}

\title{Self-Supervised Learning for Place Representation Generalization across Appearance Changes}

\author{Mohamed Adel Musallam\\
{\tt\small mohamed.ali@uni.lu}
\and Vincent Gaudilli\`ere\\
{\tt\small vincent.gaudilliere@uni.lu}
\and Djamila Aouada\\
{\tt\small djamila.aouada@uni.lu}
\and
SnT, University of Luxembourg
}

\maketitle

\begin{abstract}
   Visual place recognition is a key to unlocking spatial navigation for animals, humans and robots. While state-of-the-art approaches are trained in a supervised manner and therefore hardly capture the information needed for generalizing to unusual conditions, we argue that self-supervised learning may help abstracting the place representation so that it can be foreseen, irrespective of the conditions. More precisely, in this paper, we investigate learning features that are robust to appearance modifications while sensitive to geometric transformations in a self-supervised manner. This dual-purpose training is made possible by combining the two self-supervision main paradigms, \textit{i.e.} contrastive and predictive learning. Our results on standard benchmarks reveal that jointly learning such appearance-robust and geometry-sensitive image descriptors leads to competitive visual place recognition results across adverse seasonal and illumination conditions, without requiring any human-annotated labels.\footnote{This work was funded  by  the Luxembourg National  Research  Fund (FNR), under the project reference BRIDGES2020/IS/14755859/MEET-A/Aouada, and by LMO (https://www.lmo.space).}.
\end{abstract}

\section{Introduction}
\label{sec:intro}

\begin{figure}[h]
    \centering
    \includegraphics[width=\linewidth, trim=100mm 0mm 95mm 0mm, clip]{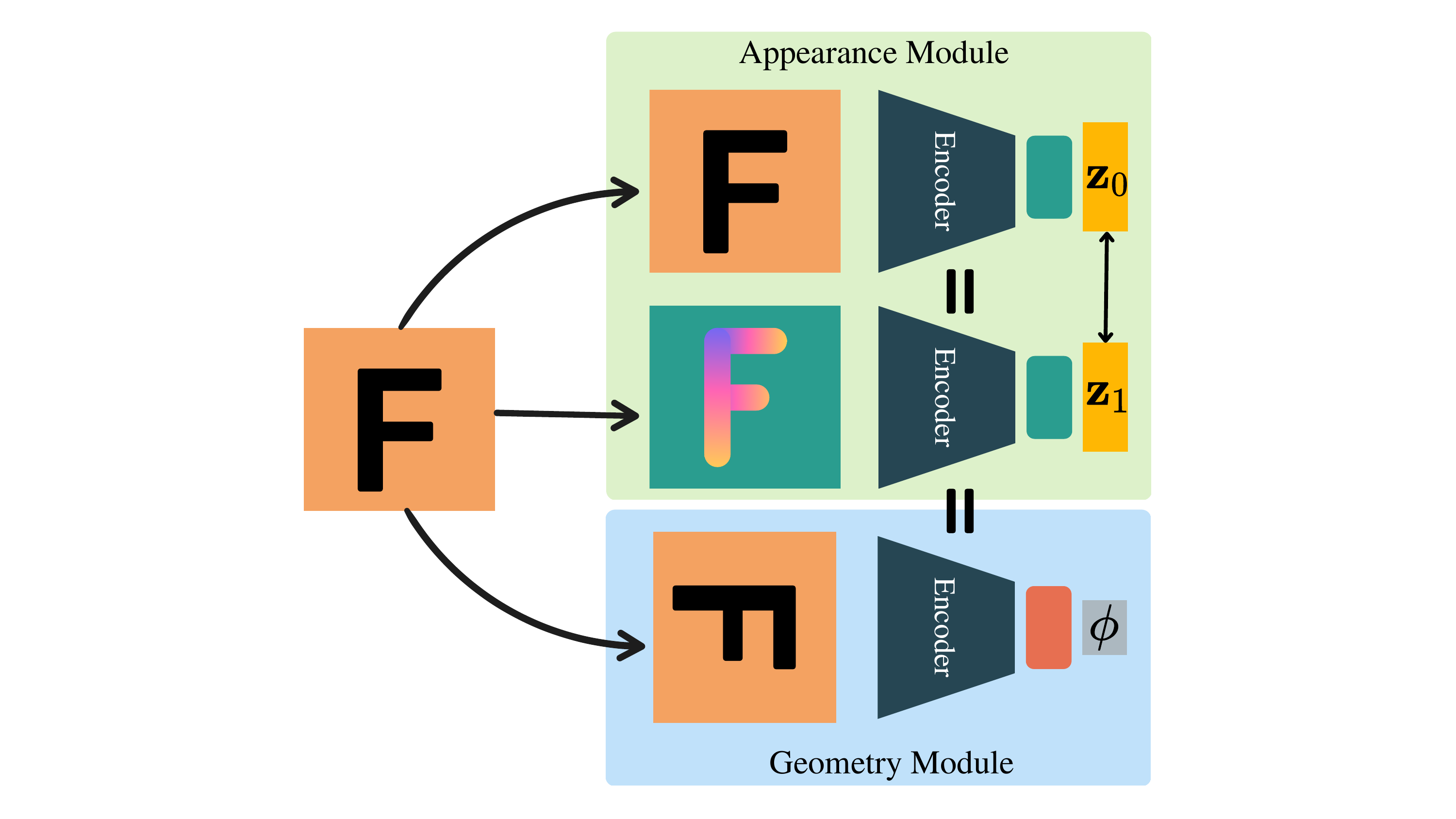}
    \caption{\textbf{CLASP-Net Training Strategy:} Three views are generated from an input image. The Appearance Module in green (top) maps the original and appearance-augmented views into close representation vectors $\left\{\mathbf{z}_{0},\mathbf{z}_{1}\right\}$. The Geometry Module in blue (bottom) predicts the transformation $\phi$ applied between the original and third views.}
    \label{fig:approach}
\end{figure}


Visual Place Recognition (VPR) is central for localizing - \textit{i.e.}  determining a camera's position in a scene~\cite{PionHCCS20,HumenbergerCPWL22}, and has applications from autonomous driving to augmented reality. Typically viewed as an image retrieval task, VPR aims to match a \textit{query} image to images in a \textit{reference}  database that depict the same location, even when conditions like viewpoint, obstructions, or weather vary~\cite{lowry2015visual}. This makes VPR challenging but vital for dependable real-world vision-based systems.


For this goal, neuroscience research indicates that biological intelligence relies on creating abstract representations of places, known as \textit{cognitive maps}~\cite{lowry2015visual}, to recognize them under varying conditions ~\cite{Whittington2022}. 

These maps are essential for generalizing limited knowledge, such as recognizing a place seen only in daylight during nighttime. The aim is to build rich representations reflecting the intrinsic structures that are not required to be re-learned from scratch when non-critical visual information changes~\cite{Whittington2022}.

In the context of technological solutions, state-of-the-art VPR methods have focused on achieving invariance to both environmental conditions and viewpoint changes in image representations. The latter are for recognizing places observed under unprecedented angles~\cite{Garg0M21,8258890}. However, we argue that such viewpoint invariance may be detrimental in the process of distinguishing between different places. Moreover, recent works have shown that favouring a more general equivariance in image representations may be more beneficial than seeking \textit{only} invariance~\cite{wang2021residual,dangovski2022equivariant,musallam2022leveraging}.


Motivated by this, we introduce \textbf{CLASP-Net }:\textit{Contrastive Learning with Appearance Augmentations and Spatial Predictions for Place Recognition}\footnote{Clasp: [noun] a device, usually of metal, for fastening together two or more things or parts of the same thing.}, designed to learn discriminative place representations that can generalize to new conditions. We use Self-Supervised Learning (SSL) to address training limitations due to low appearance variability in reference images. Contrastive Learning (CL) is employed to unify representations of the same place by using appearance augmentations. To further exploit the scene's spatial layout and regularize the model, we apply geometric transformations and utilize a Predictive Learning (PL) framework for classification based on these transformations.

Contrary to supervised learning, which tends to learn shortcuts and struggles to generalize from limited labelled data~\cite{Geirhos2020}, SSL seems closer to human-like learning and does not require manual annotation~\cite{SSL-blog-post}. Few studies have explored SSL for VPR~\cite{GargVM21,disentanglement}. We propose to merge the key SSL approaches, CL \cite{ChenK0H20} and PL \cite{SSL_survey}, aiming for image representations that are resilient to appearance shifts while being sensitive to geometric cues. By doing that, we aim to learn features suitable for visual place recognition under appearance changes.


\noindent\textbf{Contributions.} Our contributions are two-fold:

\noindent(1) A novel approach for Visual Place Recognition under extreme condition changes, CLASP-Net, that leverages both contrastive and predictive self-supervised learning approaches.\\
(2) An experimental evaluation confirming the competitiveness of CLASP-Net compared to state-of-the-art approaches on standard benchmarks featuring different conditions (day/night, weather, seasons), among which the very challenging Alderley Dataset~\cite{milford2012seqslam}.
\\
\noindent\textbf{Paper organization.} The rest of the paper is organized as follows. Relevant work on SSL and VPR is reviewed in Section \ref{sec:related}. CLASP-Net is presented in Section \ref{sec:method}, while experimental evaluation demonstrating the validity of our approach is reported in Section \ref{sec:results}. Section \ref{sec:conclusions} concludes the paper and presents future works.

\section{Related Work}
\label{sec:related}

\subsection{CNN-based Descriptors for Visual Place Recognition}

The rapid evolution of deep learning has opened new avenues for overcoming the limitations of traditional, handcrafted descriptors. Following the groundbreaking work by Chen \etal \cite{chen2014convolutional}, there has been a growing emphasis on learning-based descriptors primarily built from Convolutional Neural Networks (CNNs).
For instance, Sunderhauf \etal~\cite{sunderhauf2015performance} and Hou \etal~
\cite{hou2015convolutional} found that mid-level features from trained CNN model are more resilient to variations in appearance.

Moreover, a concerted effort has been made to design specialized neural networks for VPR tasks. This has led to the invention of techniques like CALC~\cite{merrill2018lightweight}, NetVLAD~\cite{arandjelovic2016netvlad}, NetBoW~\cite{ong2018deep} and NetFV~\cite{miech2017learnable} that meld the best aspects of both traditional and learning-based descriptors, achieving unprecedented results.

In terms of performance, CNN-based descriptors, particularly those relying on supervised learning, are highly dependent on extensive, high-quality training datasets. 

However, it's crucial to acknowledge that supervised learning methods often require laborious data annotation, which can be both time-consuming and costly. 
Therefore, self-supervised learning presents a compelling alternative to VPR tasks.

\subsection{Self-Supervised Learning}
\label{ssec:SSL}


Self-supervised methods focus on learning visual features from large sets of unlabeled images, making them valuable for diverse real-world applications such as autonomous driving. These methods usually employ a pretext task with a related objective function for training~\cite{SSL_survey}. The objective function can target either network predictions (predictive learning) or the feature representation space (contrastive learning). This enables SSL to yield image representations that are both sensitive and robust to specific transformations.


\paragraph{Predictive Learning.} PL uses pretext tasks to indirectly infuse image representations with inductive biases via network outputs~\cite{SSL_survey}. Tasks range from image colorization~\cite{ZhangIE16} and jigsaw puzzles~\cite{NorooziF16} to rotation prediction~\cite{gidaris2018unsupervised}. These tasks encourage the network to learn rich object representations and their spatial arrangements. For instance, predicting an outdoor scene often involves recognizing sky and trees at the top and roads at the bottom, requiring an understanding of the scene's structure.


\paragraph{Contrastive Learning.} CL directly refines image representations using a contrastive loss that considers batch elements' relationships. SimCLR~\cite{ChenK0H20}, a framework for visual representation through CL, stands out for its simplicity, not needing specialized structures~\cite{NEURIPS2019_ddf35421} or memory banks~\cite{Wu_2018_CVPR,Misra_2020_CVPR}. It works by sampling two distinct augmentations, applying each to an image, and then training encoders on a contrastive loss to maximize similarity between the two views and minimize similarity with different images. To address potential training convergence challenges in CL, ScatSimCLR~\cite{Kinakh_2021_ICCV} also estimates each view's augmentation parameters.

\paragraph{Combining Predictive and Contrastive Learning.}

CL aims at inducing invariance to some content-preserving transformations while being distinctive to such content changes. On the other side, PL is mostly used to incorporate sensitivity, and ideally equivariance, to given transformations into representations~\cite{wang2021residual}. Some studies have demonstrated the advantage of balancing invariance and equivariance~\cite{Patrick_2021_ICCV,WangSPJZ22,garrido2023self}. For example, Winter et al.\cite{WinterBLNC22} suggested an AutoEncoder-centric framework to cultivate representations that exhibit both robustness and sensitivity to rotations. Explicitly, an encoder translates a rotated image into a more invariant latent representation, from which a decoder predicts the unrotated original image. Simultaneously, an auxiliary branch pursues equivariance by determining the rotation angle. In a similar vein, Feng \etal~\cite{Feng_2019_CVPR} endeavour to learn features impervious to the rotation of input images by bifurcating the features: one segment is dedicated to rotation prediction (dubbed equivariant features), while another segment, subjected to a contrastive loss, penalizes disparities emerging from various rotations (termed invariant features).

In recent work, Dangovski \etal~\cite{dangovski2022equivariant} introduced Equivariant Self-Supervised Learning (E-SSL), a more nuanced SSL approach that goes beyond simply seeking invariant representations. E-SSL framework enriches traditional SSL methods by integrating both equivariance and invariance objectives in the pre-training process. The key insight is that some transformations are better captured as equivariant, meaning that the learned features should change predictably based on how the input is transformed. At the same time, other transformations are better captured as invariant, where the feature representation should remain constant despite changes of the input.


Drawing on these insights, our proposed CLASP-Net focuses on achieving appearance invariance through CL while capturing detailed representations of scene components and their spatial layouts through PL. With the latter, the network gains sensitivity to geometric transformations, enhancing its suitability for VPR tasks.

\subsection{Self-Supervised Learning for Visual Place Recognition}


As highlighted in Section \ref{ssec:SSL}, SSL is well-suited for VPR because it addresses the issue of unrepresentative training data due to varying test conditions. Despite its promise, few methods exist. For instance, Tang \etal~\cite{disentanglement} have proposed to disentangle appearance-related and place-related features using a generative adversarial network with two discriminators. However, this type of method may suffer from unstable training. SeqMatchNet~\cite{GargVM21} is a CL-based method that leverages sequences of video frames in the contrastive loss to robustify image representations for VPR.


From a larger perspective, Mithun et al.\cite{MithunSCHR18} use sets of related images (i.e., showing the same place under different conditions) to enhance VPR image representations. Thoma et al.\cite{NEURIPS2020_7f2cba89} suggest loosening geo-tag constraints for weakly-supervised training. Unlike these works, we generate pairs of corresponding images in a self-supervised manner, without labels. Venator et al.~\cite{VenatorHABM21} employ SSL to create appearance-invariant descriptors for image matching, which could serve as a refinement step in our approach.

\section{Proposed CLASP-Net}
\label{sec:method}

\begin{figure*}
    \centering
    \includegraphics[trim=0mm 50mm 0mm 60mm, clip, width=\linewidth]{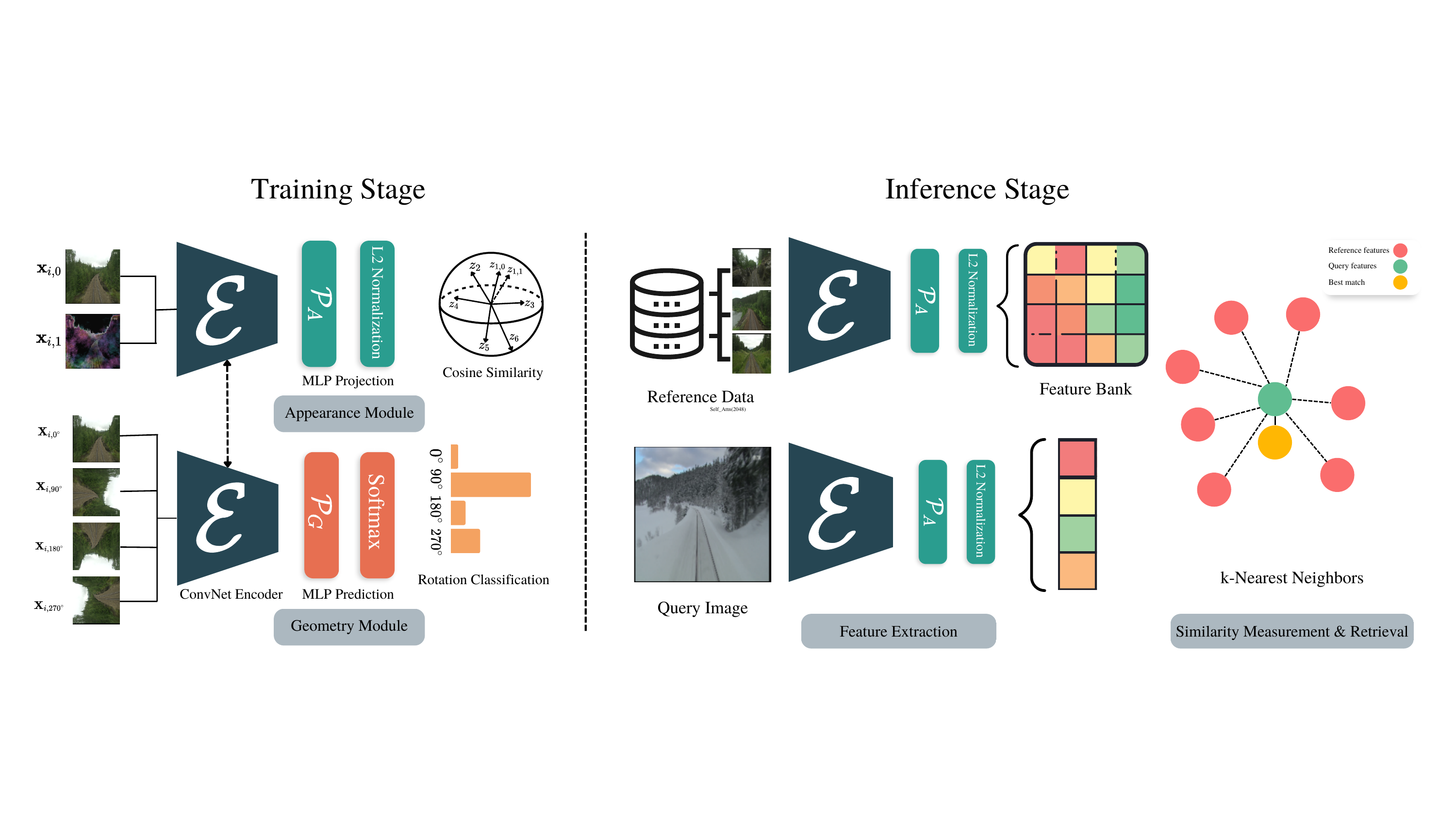}
    \caption{Overview of CLASP-Net. \textbf{Training Stage:} from an original image $\im_{i,0}$, augmented versions with a modified appearance $\im_{i,1}$ and different orientations $(\im_{i,0\degree}, \im_{i,90\degree}$, $\im_{i,180\degree}$, $\im_{i,270\degree})$ are generated. Representations of the first two images are brought closer thanks to a contrastive learning framework to achieve appearance robustness. In parallel, original and rotated images are passed through a classification network sharing the same encoder to predict the applied transformation and achieve geometric sensitivity. Note that our method does not rely on any manual annotation. \textbf{Inference Stage:} The representations from query and reference images are compared based on similarity measure then the closest $k$ reference images constitute the image retrieval output.}
    \label{fig:full_model}
\end{figure*}


Our primary objective is to enable the model to learn features that can withstand drastic changes in appearance while remaining effective for VPR. Specifically, we aim to create image representations that capture essential geometric details of the scene's spatial arrangement yet remain unaffected by varying environmental conditions. To accomplish this, we integrate both sensitivity to geometric information and robustness to appearance changes into the image representations using self-supervised learning techniques.

\subsection{Problem Formalization}

Following the traditional approach~\cite{lowry2015visual}, we frame the VPR problem as an image retrieval task, where, given a query image $\mathbf{q}$ depicting a place $\mathscr{P}_\mathbf{q}$, a representation \textit{a.k.a.} descriptor $\mathbf{z_q}$ of that image is computed. It is then compared to the descriptors $\{\desc_i\}_{i=1..N_R}$ of reference images $\{\im_i\}_{i=1..N_R}$, where $N_R$ is the size of the reference database. The comparison is done using a given similarity metric (\textit{e.g.}, cosine similarity). This inference stage is illustrated in Figure~\ref{fig:full_model}.

During the training, the model only has access to reference images that we assume unlabelled. Moreover, the environmental conditions under which the query image is acquired are not necessarily similar to the ones featured in the reference database, making the problem very challenging, even sometimes for human eyes.

\subsection{Preliminaries: Robustness \& Sensitivity}
\label{ssec:inveq}

Our approach focuses on extracting image features that are both robust to appearance changes and sensitive to geometric aspects. Mathematically, these properties correspond to the concepts of invariance and equivariance. Formally, let \( \mathfrak{G} \) be a generic group of transformations and \( \mathfrak{g} \) an element of \( \mathfrak{G} \). The actions of \( \mathfrak{g} \) on the input and output spaces of a function \( \mathcal{F}:\mathbb{I}\rightarrow\mathbb{O} \) are denoted by \( \phi_\mathfrak{g}^{(\mathbb{I})} \) and \( \phi_\mathfrak{g}^{(\mathbb{O})} \), respectively.






In practice, considering an encoder model $\encoder$ for extracting features from an image $\im$, we seek robustness to any appearance transformation $\mathcal{T}_A$:
\begin{equation}\label{eq:invariance2}
    \forall \mathcal{T}_A, \forall i\in [1;N_R], \quad \encoder(\mathcal{T}_A\im_i)\approx\encoder(\im_i)\mbox{,}
\end{equation}
while, at the same time, sensitivity to a certain group of geometric transformations $\mathfrak{G}_G$:
\begin{equation}\label{eq:invariance2}
    \forall \mathcal{T}_G\in\mathfrak{G}_G, \forall i\in [1;N_R], \quad \encoder(\mathcal{T}_G\im_i)\approx\mathcal{T}_G'\encoder(\im_i)\mbox{,}
\end{equation}
where $\mathcal{T}_G'\approx\mathcal{T}_G$.
The different possible groups of transformations are investigated in Section \ref{sec:results}.

\subsection{Model Architecture}


Our pipeline exploits both CL for encouraging invariance to appearance changes and PL for encouraging sensitivity to geometric image augmentations. This hybrid approach is consistent with the \textit{E-SSL} framework proposed in~\cite{dangovski2022equivariant}. 
The overall architecture of the proposed CLASP-Net is presented in Figure \ref{fig:full_model}.

At training time, CLASP-Net is composed of two branches sharing the weights of an encoder model $\encoder$. The first branch, denoted \textit{Appearance Module}, takes as inputs the original image $\im_i$ and an augmented version with modified appearance $\mathcal{T}_A\im_i$, then applies a contrastive learning loss in the representation space to bring the two descriptors closer. The second branch, denoted \textit{Geometry Module}, uses rotated versions of the original image, $\mathcal{T}_G'\im_i = \mathrm{R}(n\degree)\im_i$, and predicts the angle of the rotation $n$. 




\noindent\textbf{Appearance Module.} The first branch, divided into two sub-branches (see Figure~\ref{fig:approach}), This setup is inspired by SimCLR \cite{ChenK0H20} and employs a shared encoder $\encoder$ and MultiLayer Perceptron (MLP) $\MLPA$ mapping between the image domain and the latent representation space where the contrastive loss is applied. Given original images $\im_i$ along with their augmented versions $\mathcal{T}_A\im_i$, the weights of the two networks are learned using a contrastive loss. This loss, formalized in Section~\ref{ssec:loss}, ensures that the descriptor of each version, \textit{e.g.}, $\encoder(\MLPA(\im_i))$, is similar to the descriptor of its corresponding view, $\encoder(\MLPA(\mathcal{T}_A\im_i))$, while distant from the other descriptors. The intuition behind this module is to force the encoder model $\encoder$ to learn features agnostic on the conditions (e.g. illumination, weather, season) under which the place was initially observed.



\noindent\textbf{Geometry Module.} The second branch incorporates the same shared encoder $\encoder$ along with a prediction-focused $\MLPG$. This setup is designed to classify rotated versions of the original image, denoted  $\mathrm{R}(n\degree)\im$, based on their rotation angle $n$. Utilizing a standard cross-entropy loss, the module aims to train the encoder $\encoder$ to learn rich representations of scene layout and spatial arrangement and capture geometry-sensitive features vital for accurate place recognition.

Combined, these two modules work together to disentangle appearance and geometric aspects of input images, enabling robust visual place recognition even when appearance conditions vary. During inference, the architecture used to compute image descriptors consists of the encoder $\encoder$  followed by the projector network $\MLPA$ , as shown in Figure \ref{fig:full_model} (right part).

\subsection{Model Loss}
\label{ssec:loss}

\begin{figure*}[ht]
    \centering
    \includegraphics[trim=24mm 24mm 24mm 24mm, clip, width=\linewidth]{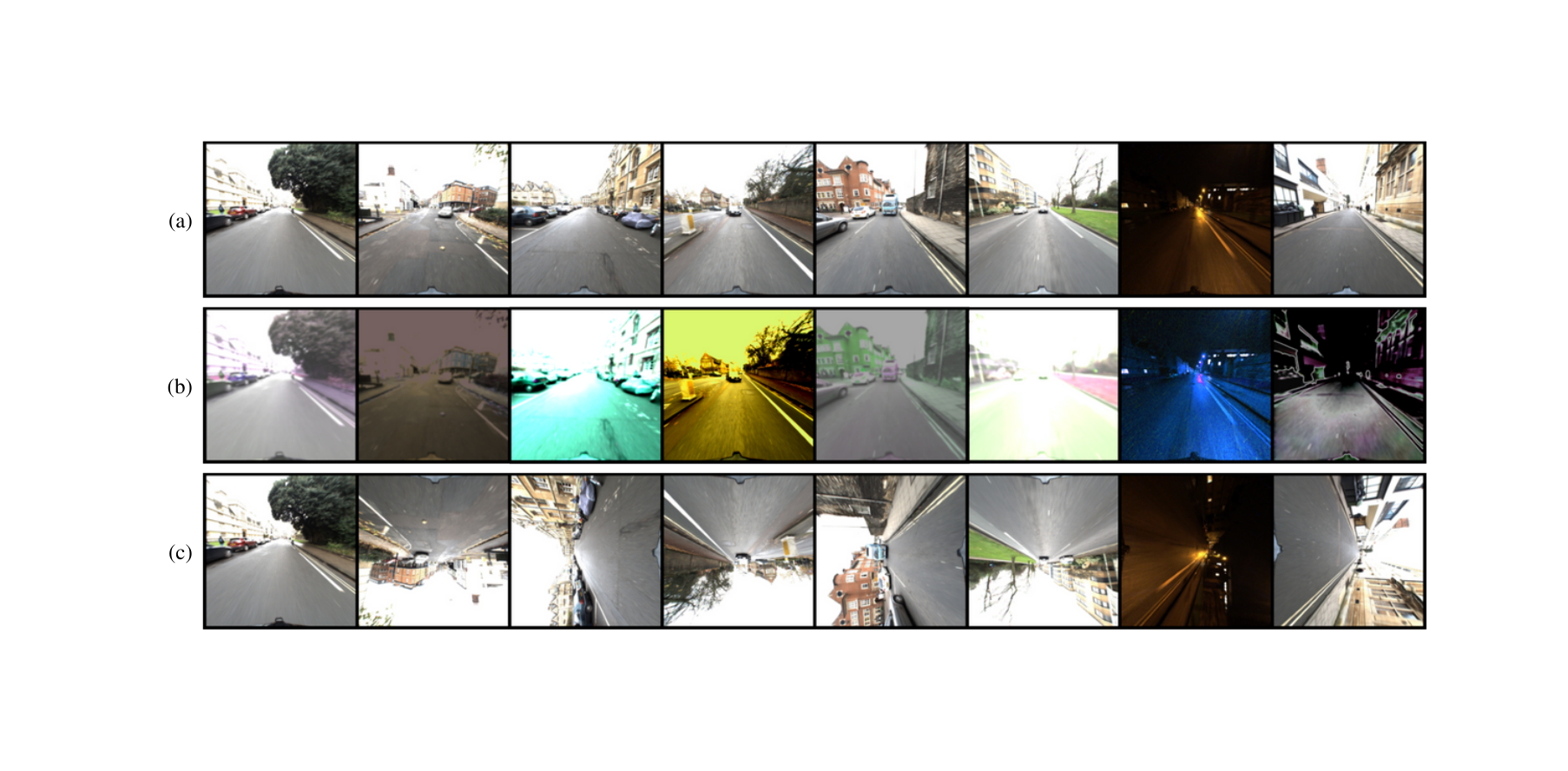}
    \caption{Examples of augmentations leveraged by CLASP-Net. Top row (a): an original input batch from Oxford RobotCar v2 dataset, (b) pixel-level augmentations for appearance changes, (c) examples of rotations applied to the original image.}
    \label{fig:batch_aug}
\end{figure*}

\noindent\textit{Note: For the sake of clarity, we herein introduce more specific notations for denoting images and their augmented/rotated versions.}

We use a combination of contrastive and predictive losses to steer our model toward robustness to appearance changes and sensitivity to geometric variations.


Given a random batch of $N$ reference images $\mathcal{B}=\{\im_{i,0}\}_{i=1..N}$ corresponding to $N$ different places, we apply one random appearance transformation to each image. By so doing, we create $N$ additional images $\{\im_{i,1}\}_{i=1..N}$. These $2N$ images constitute the contrastive batch $\mathcal{B}_C=\{\im_{i,j}\}_{i=1..N, j\in \{0,1\}}$ that is fed into the Appearance Module. Furthermore, we also apply rotations of $0\degree$, $90\degree$, $180\degree$ and $270\degree$ to each original image. As a result, we create the predictive batch of $4N$ images $\mathcal{B}_P=\{\im_{i,j\degree}\}_{i=1..N, j\in\Theta_4}$, where $\Theta_4=\{0,90,180,270\}$. $\mathcal{B}_P$ is fed into the Geometry  Module.


\paragraph{Contrastive loss.} The contrastive batch $\mathcal{B}_C$ contains $N$ \textit{positive} pairs of images $(\im_{i,0},\im_{i,1})$ depicting the same place, the rest being \textit{negative} pairs corresponding to different places. We use NT-Xent loss~\cite{ChenK0H20} that leverages positive samples, and is based on the cosine similarities between the obtained image representations $\desc_{.,.}=\MLPA(\encoder(\im_{.,.}))$, expressed as
\begin{equation}
    \mathrm{s}(\desc_{i,j},\desc_{k,l})=\frac{\desc_{i,j}\cdot\desc_{k,l}}{\|\desc_{i,j}\| \|\desc_{k,l}\|}\mbox{,}
\end{equation}
where $\cdot$ is the dot product.
\\
Specifically, the contrastive loss is defined as
\begin{equation}
    \mathcal{L}_C = \frac{1}{2N}\sum_{i=1}^{N}\ell_{0\rightarrow 1}(i)+\ell_{1\rightarrow 0}(i)\mbox{,}
\end{equation}
where
\begin{equation}
    \ell_{a\rightarrow b}(i)=-\mathrm{log}\frac{\mathrm{exp}(\mathrm{s}(\desc_{i,a},\desc_{i,b})/\tau)}{\sum_{k=1}^{N}\mathds{1}_{k\neq i}\sum_{j=0}^{1}\mathrm{exp}(\mathrm{s}(\desc_{i,a},\desc_{k,j})/\tau)}\mbox{,}
\end{equation}
with $\tau$ denoting a temperature parameter that controls the strength of penalties on pairs of non-corresponding images~\cite{Wang_2021_CVPR} and $\mathds{1}_{k\neq i}$ being equal to 1 if $k\neq i$, and 0 otherwise.

The contrastive loss aims at making representations of the same place under different conditions similar to each other, while forcing representations of different places to be different.

\noindent\textbf{Predictive loss.} The predictive batch $\mathcal{B_P}$ contains four rotated views of each place. The task of this branch is to predict the rotation angle for each of the $4N$ pictures. We frame this as a classification problem with 4 classes corresponding to $0\degree$, $90\degree$, $180\degree$ and $270\degree$ rotation angles. The predictive loss is therefore the standard cross-entropy loss:
\begin{equation}
    \mathcal{L}_P = -\sum_{i=1}^{N}\sum_{j\in\Theta_4}\mathrm{c}(\im_{i,j})\cdot\mathrm{log}(\widetilde{\desc}_{i,j})\mbox{,}
\end{equation}
where $\widetilde{\desc}_{i,j}=\mathrm{Softmax}(\MLPG(\encoder(\im_{i,j})))\in\mathbb{R}^4$ is the prediction, $\mathrm{log}()$ the element-wise natural logarithm, $\cdot$ the dot product and $\mathrm{c}(\im_{i,j})\in\mathbb{R}^4$ the groundtruth with elements equal to 0 except the $n$th element equal to 1 if the true rotation is $(n-1)\times90$\degree.

\paragraph{Overall loss.} The final loss is the combination of the contrastive loss for appearance robustness and predictive loss for geometry sensitivity:
\begin{equation}
    \mathcal{L}=\mathcal{L}_C+\lambda.\mathcal{L}_P\mbox{,}
\end{equation}
where $\lambda$ is a weighting factor to balance the two terms.

\section{Experimental Evaluation}
\label{sec:results}


\begin{table}[t]
\centering
\begin{tabular}{c|c}
\hline
\rowcolor[HTML]{C0C0C0} 
Data Augmentation Type & Probability \\ \hline
Planckian Jitter       & 0.8         \\
\rowcolor[HTML]{EFEFEF} 
Color Jiggle           & 0.5         \\
Plasma Brightness      & 0.5         \\
\rowcolor[HTML]{EFEFEF} 
Plasma Contrast        & 0.3         \\
Gray scale             & 0.3         \\
\rowcolor[HTML]{EFEFEF} 
Box Blur               & 0.5         \\
Channel Shuffle        & 0.5         \\
\rowcolor[HTML]{EFEFEF} 
Motion Blur            & 0.3         \\
Solarize               & 0.5         \\ \hline
\end{tabular}
\caption{List of data augmentations applied to the images on-the-fly during training. We also set a probability for each one of them.}
\label{tab:augmentations}

\end{table}


\subsection{Datasets}
\label{ssec:dataset}


\textbf{The Nordland dataset~\cite{sunderhauf2013we}:} records a 728 km long train journey connecting the cities of Trondheim and Bodø in Norway. It contains four long traversals, once per season, with diverse visual conditions. The dataset has 35768 images per season with one-to-one correspondences between them. We follow the dataset partition proposed by Olid \etal~\cite{olid2018single} with test set made of 3450 photos from each season.

\textbf{The Alderley dataset~\cite{milford2012seqslam}:} records an 8 km travel along the suburb of Alderley in Brisbane, Australia. The dataset contains two sequences: the first one was recorded during a clear morning, while the second one was collected on a stormy night with low visibility, which makes it a very challenging benchmark. The dataset contains 14607 images for each sequence and each place have 2 images. We train our approach on the day sequence and test on the night sequence.



\textbf{The Oxford RobotCar Seasons v2 dataset~\cite{toft2020long}:} is based on the RobotCar dataset~\cite{RobotCarDatasetIJRR}, which depicts the city of Oxford, UK. It contains images acquired from three cameras mounted on a car. There are 10 sequences corresponding to 10 different traversals carried out under very different weather and seasonal conditions. The rear camera images of the \textit{overcast-reference} traversal (6954 images) are used as a basis for reference training images, to which we add 1906 rear camera images from other traversals following the \textit{v2} train/test split. These additional images cover different environmental conditions but only a subset of places (not full traversals). The test set contains 1872 images from all traversals except \textit{overcast-reference}, without overlap with training images.

\subsection{Evaluation}
\label{ssec:eval}

\begin{table}[]
\centering
\begin{tabular}{l|ccc}
\hline
\multirow{2}{*}{Method} & \multicolumn{3}{c}{Nordland Summer/Winter}        \\ \cline{2-4} 
                        & R@1 & R@5 & R@10 \\ \hline
NetVLAD \cite{arandjelovic2016netvlad}                 & 7.7          & 13.7        & 17.7        \\
SFRS \cite{ge2020self}                    & 18.8         & 32.8        & 39.8        \\
SuperGlue \cite{sarlin2020superglue}              & 29.1         & 33.5        & 34.3        \\
DELG  \cite{cao2020unifying}                  & 51.3         & 66.8        & 69.8        \\
Patch-NetVLAD   \cite{Hausler_2021_CVPR}        & 46.4         & 58.0        & 60.4        \\
TransVPR  \cite{Wang_2022_CVPR}             & \textbf{58.8 }   & \textbf{75.0 }  & \textit{78.7}        \\
\hline
CLASP-Net (Ours)          & \textit{53.0}         & \textit{73.8}        & \textbf{80.2}        \\ \hline
\end{tabular}
\caption{Quantitative results on Nordland dataset. Best results are in \textbf{bold}. Second best results are in \textit{italic}.}
\label{tab:nordland_results}
\end{table}

\begin{table}[]
\centering
\begin{tabular}{l|c}
\hline
\multicolumn{1}{c|}{Method} & Alderley Day/Night \\ \hline
NetVLAD   \cite{arandjelovic2016netvlad}                   & 3.35               \\
CIM \cite{facil2019condition}                     & 7.82               \\
Patch-NetVLAD   \cite{Hausler_2021_CVPR}                   & 7.99              \\
Seqslam \cite{milford2012seqslam}                     & 9.90               \\
Retrained NetVLAD   \cite{tang2020adversarial}                   & 15.8               \\
AFD  \cite{tang2020adversarial}                       & 21.0               \\
\hline
CLASP-Net (Ours)              & \textbf{25.2}               \\ \hline
\end{tabular}
\caption{Quantitative results on Alderley dataset. Best result is in \textbf{bold}.}
\label{tab:alderley_results}
\end{table}

The evaluation on both Nordland and Alderley datasets uses the recall R@N measure, which consists in the proportion of successfully localized query images when considering the first $N$ retrievals. If at least one of the top $N$ reference images is within a tolerance window around the query's ground truth correspondence, the query image is deemed succesfully localized.
The tolerance window is set to two frames distant from the query before and after, so that the window contains 5 pictures. Following the common approach for NordLand~\cite{camara2020visual,hausler2020hierarchical, Hausler_2021_CVPR}, images of the winter sequence are used as queries, while the summer sequence is used as reference.

For RobotCar-Seasons v2, we follow the Patch-NetVLAD~\cite{Hausler_2021_CVPR} approach and utilize the 6-DoF pose of the best-matched reference picture as prediction of the query's pose. Since we don't compute any pose, our image retrieval method is not comparable with pose estimation methods such as MegLOC~\cite{abs-2111-13063}. 

\subsection{Implementation details}
\label{ssec:impl}

\paragraph{Encoder model $\encoder$.} We use ResNet50~\cite{ResNet} as the backbone, with pre-training on ImageNet using the Timm library \cite{rw2019timm}. The last classification layer is discarded so that the model is only used for the feature extraction.

\paragraph{Rotation predictor $\MLPG$.} We use a simple 1-layer perceptron with layer normalization and ReLU activation.

\paragraph{Projector $\MLPA$} We use a simple 1-layer perceptron with batch normalizations and ReLu activation. The dimension of the output (\textit{i.e.}, image descriptor) is 1024.

\paragraph{Appearance Augmentations.} Following domain generalization approaches, our model leverages numerous pixel-level data augmentations to trigger appearance invariance bias in the model. The list of pixel-level augmentations for appearance modification is provided in Table~\ref{tab:augmentations}, while examples of such augmentations are provided in Figure~\ref{fig:batch_aug}. The chosen set of variations empirically achieved good performance whereas other tested combinations were less favourable. We use the Kornia~\cite{eriba2019kornia} library for self-supervised data augmentation.

\begin{figure}
    \centering
    \includegraphics[trim=0mm 2mm 0mm 0mm, clip, width=.9\linewidth]{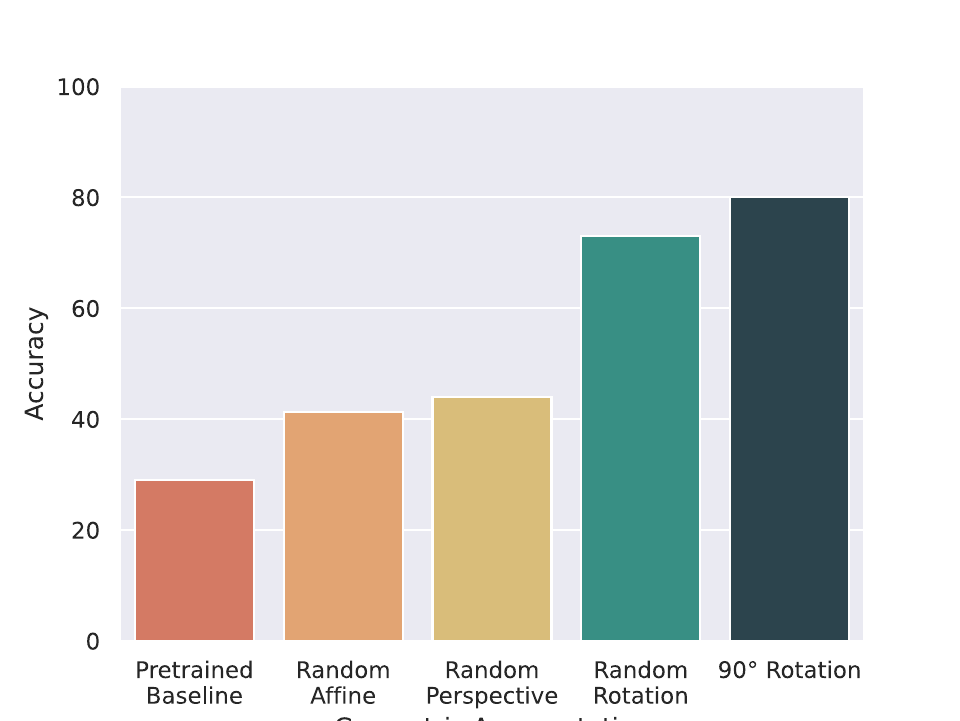}
    \caption{R@10 on Nordland Summer/Winter dataset with Geometry Modules relying on different groups of transformations.}
    \label{fig:geom_abl}
\end{figure}

\paragraph{Geometric Augmentations.} Our training strategy encourages information about rotations to be retained in the image representation rather than guaranteeing strict equivariance. 
Moreover, the choice of this particular group of geometric transformations is the outcome of experimentations whose results are presented in Figure \ref{fig:geom_abl}. Empirically, we found that the best performance is achieved with the cyclic group of 90$\degree$ rotations, compared to the groups of 2D affine transformations, 2D projective transformations, and 2D rotations.

\paragraph{Model training.} The model is trained for 1000 epochs using Adam optimizer~\cite{kingma2014adam} and a batch size of 64. Although contrastive learning usually requires larger batch size~\cite{chen2020simple}, using Adam optimizer allowed us to obtain good results with a smaller batch size. A learning rate of 0.003 had the best performance with this optimizer. The temperature parameter $\tau$ is set to 0.01 and the loss factor $\lambda$ is set to 1 in our experiments. 

\paragraph{Inference.} Prior to the inference stage, we pass the set of reference images to the Appearance Invariant Module of the trained model: $\encoder\rightarrow\MLPA\rightarrow L2\mathrm{-normalization}$ and thus build a reference descriptor bank. A k-Nearest Neighbor search based on cosine similarity to find the closest references to the query image.



\subsection{Results}
\label{ssec:res}

\begin{table*}[!ht]
\centering
\resizebox{\textwidth}{!}{%

\begin{tabular}{l|ccccccc|cc}
 &
  \multicolumn{7}{c|}{day conditions} &
  \multicolumn{2}{c}{night conditions} \\ \cline{2-10} 
 &
  \multicolumn{1}{c|}{dawn} &
  \multicolumn{1}{c|}{dusk} &
  \multicolumn{1}{c|}{OC-summer} &
  \multicolumn{1}{c|}{OC-winter} &
  \multicolumn{1}{c|}{rain} &
  \multicolumn{1}{c|}{snow} &
  sun &
  \multicolumn{1}{c|}{night} &
  night-rain \\ \cline{2-10} 
\multicolumn{1}{r|}{m} &
  \multicolumn{1}{c|}{.25 / .50 / 5.0} &
  \multicolumn{1}{c|}{.25 / .50 / 5.0} &
  \multicolumn{1}{c|}{.25 / .50 / 5.0} &
  \multicolumn{1}{c|}{.25 / .50 / 5.0} &
  \multicolumn{1}{c|}{.25 / .50 / 5.0} &
  \multicolumn{1}{c|}{.25 / .50 / 5.0} &
  .25 / .50 / 5.0 &
  \multicolumn{1}{c|}{.25 / .50 / 5.0} &
  .25 / .50 / 5.0 \\
\multicolumn{1}{r|}{deg} &
  \multicolumn{1}{c|}{2 / 5 / 10} &
  \multicolumn{1}{c|}{2 / 5 / 10} &
  \multicolumn{1}{c|}{2 / 5 / 10} &
  \multicolumn{1}{c|}{2 / 5 / 10} &
  \multicolumn{1}{c|}{2 / 5 / 10} &
  \multicolumn{1}{c|}{2 / 5 / 10} &
  2 / 5 / 10 &
  \multicolumn{1}{c|}{2 / 5 / 10} &
  2 / 5 / 10 \\ \hline
AP-GEM \cite{revaud2019learning} &
  \multicolumn{1}{c|}{1.4 / 14.2 / 65.9} &
  \multicolumn{1}{c|}{9.6 / 29.4 / 82.9} &
  \multicolumn{1}{c|}{2.4 / 19.1 / 80.5} &
  \multicolumn{1}{c|}{3.6 / 20.3 / 78.1} &
  \multicolumn{1}{c|}{4.4 / 21.5 / 86.0} &
  \multicolumn{1}{c|}{4.5 / 15.8 / 75.9} &
  1.8 / 7.5 / 58.2 &
  \multicolumn{1}{c|}{0.0 / 0.2 / 6.8} &
  0.1 / 1.2 / 15.8 \\
DenseVLAD \cite{torii201524} &
  \multicolumn{1}{c|}{4.5 / 24.3 / 79.6} &
  \multicolumn{1}{c|}{12.5 / 38.9 / 89.1} &
  \multicolumn{1}{c|}{3.8 / 27.4 / 90.8} &
  \multicolumn{1}{c|}{4.1 / 27.1 / 85.6} &
  \multicolumn{1}{c|}{5.4 / 29.0 / 91.4} &
  \multicolumn{1}{c|}{6.7 / 25.5 / 85.1} &
  3.2 / 11.0 / 67.1 &
  \multicolumn{1}{c|}{\textbf{1.4} / 2.7 / 23.2} &
  0.6 / 5.2 / 29.8 \\
NetVLAD \cite{arandjelovic2016netvlad}&
  \multicolumn{1}{c|}{2.2 / 16.8 / 73.3} &
  \multicolumn{1}{c|}{11.4 / 31.0 / 85.9} &
  \multicolumn{1}{c|}{3.2 / 21.5 / 90.9} &
  \multicolumn{1}{c|}{4.1 / 22.6 / 84.0} &
  \multicolumn{1}{c|}{4.2 / 22.2 / 89.4} &
  \multicolumn{1}{c|}{5.2 / 20.1 / 80.8} &
  2.4 / 10.4 / 70.3 &
  \multicolumn{1}{c|}{0.2 / 1.2 / 9.1} &
  0.3 / 0.9 / 8.8 \\
DELG global \cite{cao2020unifying} &
  \multicolumn{1}{c|}{1.6 / 10.9 / 66.4} &
  \multicolumn{1}{c|}{8.9 / 23.9 / 81.3} &
  \multicolumn{1}{c|}{2.1 / 16.5 / 77.6} &
  \multicolumn{1}{c|}{3.5 / 18.5 / 73.6} &
  \multicolumn{1}{c|}{3.9 / 20.5 / 87.9} &
  \multicolumn{1}{c|}{3.6 / 13.5 / 73.5} &
  1.0 / 6.4 / 59.6 &
  \multicolumn{1}{c|}{0.2 / 0.7 / 7.6} &
  0.1 / 1.6 / 13.8 \\
DELG local \cite{cao2020unifying} &
  \multicolumn{1}{c|}{1.7 / 10.4 / 78.3} &
  \multicolumn{1}{c|}{2.5 / 7.3 / 76.8} &
  \multicolumn{1}{c|}{1.1 / 8.9 / 84.2} &
  \multicolumn{1}{c|}{1.2 / 9.1 / 83.2} &
  \multicolumn{1}{c|}{1.2 / 4.5 / 76.8} &
  \multicolumn{1}{c|}{3.5 / 10.9 / 80.8} &
  3.3 / 12.6 / \textit{85.2} &
  \multicolumn{1}{c|}{\textbf{1.4} / \textit{7.6 }/ \textbf{38.6}} &
  \textbf{2.4 }/ \textit{11.9} / \textbf{53.0} \\
SuperGlue \cite{sarlin2020superglue}&
  \multicolumn{1}{c|}{4.3 / 24.6 / 84.8} &
  \multicolumn{1}{c|}{12.7 / 40.3 / 88.6} &
  \multicolumn{1}{c|}{5.0 / 31.5 / \textit{95.0}} &
  \multicolumn{1}{c|}{\textit{4.5} / 30.2 / 88.6} &
  \multicolumn{1}{c|}{5.9 / 30.1 / 91.8} &
  \multicolumn{1}{c|}{7.0 / 25.4 / 87.2} &
  3.3 / 17.1 / 83.9 &
  \multicolumn{1}{c|}{0.5 / 2.2 / 27.9} &
  0.9 / 5.4 / \textit{31.8} \\
Patch-NetVLAD~\cite{Hausler_2021_CVPR} &
  \multicolumn{1}{c|}{4.8 / \textbf{72.5} / 86.2} &
  \multicolumn{1}{c|}{\textbf{13.5} / \textbf{72.0} / 89.5} &
  \multicolumn{1}{c|}{5.3 / \textbf{80.9} / 94.5} &
  \multicolumn{1}{c|}{\textbf{6.3} / \textbf{71.3} / 89.8} &
  \multicolumn{1}{c|}{5.9 / \textbf{79.3} / 92.1} &
  \multicolumn{1}{c|}{7.8 / \textbf{75.9 }/ 87.9} &
  4.8 / \textbf{67.3} / 83.4 &
  \multicolumn{1}{c|}{0.5 / \textbf{12.4 }/ 24.9} &
  \textit{1.0} / \textbf{19.0 }/ 30.8 \\

TransVPR~\cite{Wang_2022_CVPR} &
  \multicolumn{1}{c|}{\textbf{18.5} / 52.0 / \textbf{95.6}} &
  \multicolumn{1}{c|}{10.7 / 44.7 / \textbf{100.0}} &
  \multicolumn{1}{c|}{\textbf{12.3} / \textit{45.5} / \textbf{99.1}} &
  \multicolumn{1}{c|}{1.2 / \textit{36.6} / \textbf{99.4}} &
  \multicolumn{1}{c|}{\textbf{15.1} / 50.7 / \textbf{99.5}} &
  \multicolumn{1}{c|}{\textbf{14.0} / \textit{42.8} / \textbf{99.1}} &
  \textbf{13.4} / \textit{34.4} / \textbf{91.1} &
  \multicolumn{1}{c|}{\textit{0.9} / 4.9 / \textit{30.5}} &
  0.0 / 1.0 / 10.3 \\ \hline
\textbf{CLASP-Net (Ours)} &
  \multicolumn{1}{c|}{8.4 / 26.9 / \textit{88.1}} &
  \multicolumn{1}{c|}{5.1 / 25.9 / \textit{89.8}} &
  \multicolumn{1}{c|}{7.1 / 32.7 / 84.4} &
  \multicolumn{1}{c|}{0.6 / 22.6 / \textit{91.5}} &
  \multicolumn{1}{c|}{\textit{12.7} / 42.9 / \textit{93.7}} &
  \multicolumn{1}{c|}{\textit{8.8} / 31.2 / \textit{90.2}} &
  \textit{8.9} / 22.3 / 76.8 &
  \multicolumn{1}{c|}{0.0 / 2.3 / 14.0} &
  0.0 / 3.0 / 14.8 \\ \hline
\end{tabular}}
\caption{Quantitative results on RobotCar Seasons v2 dataset. Best results are in   \textbf{bold}. Second best results are in \textit{italic}.}
\label{tab:robotcar_v2_results}

\end{table*}

Tables \ref{tab:nordland_results}, \ref{tab:alderley_results} and \ref{tab:robotcar_v2_results} show the results of CLASP-Net along with other approaches on the three previously described datasets: partitioned Nordland, Alderley Day/Night and RobotCar-Seasons datasets.

The results demonstrate that our method outperforms, by a large margin, standard baselines such as NetVLAD~\cite{arandjelovic2016netvlad} and even local feature-based methods such as SuperGlue~\cite{sarlin2020superglue}. It outperforms Patch-NetVLAD~\cite{Hausler_2021_CVPR} on Nordland dataset (Table \ref{tab:nordland_results}) and competes with it on Robotcar Seasons v2 (Table \ref{tab:robotcar_v2_results}), despite the fact that Patch-NetVLAD leverages multi-scale descriptors whereas we rely on a single global descriptor. Only the transformer-based architecture TransVPR~\cite{Wang_2022_CVPR} presents a higher performance as compared to CLASP-Net. We note, however, that our model is based on simple ConvNet and MLP elements that can be upgraded to improve the performance. Finally, it is worth noting that we achieve state-of-the-art results on the very challenging Alderley dataset (Table~\ref{tab:alderley_results}).

\begin{figure}
    \centering
    \includegraphics[trim=19mm 22mm 23mm 20mm, clip, width=\linewidth]{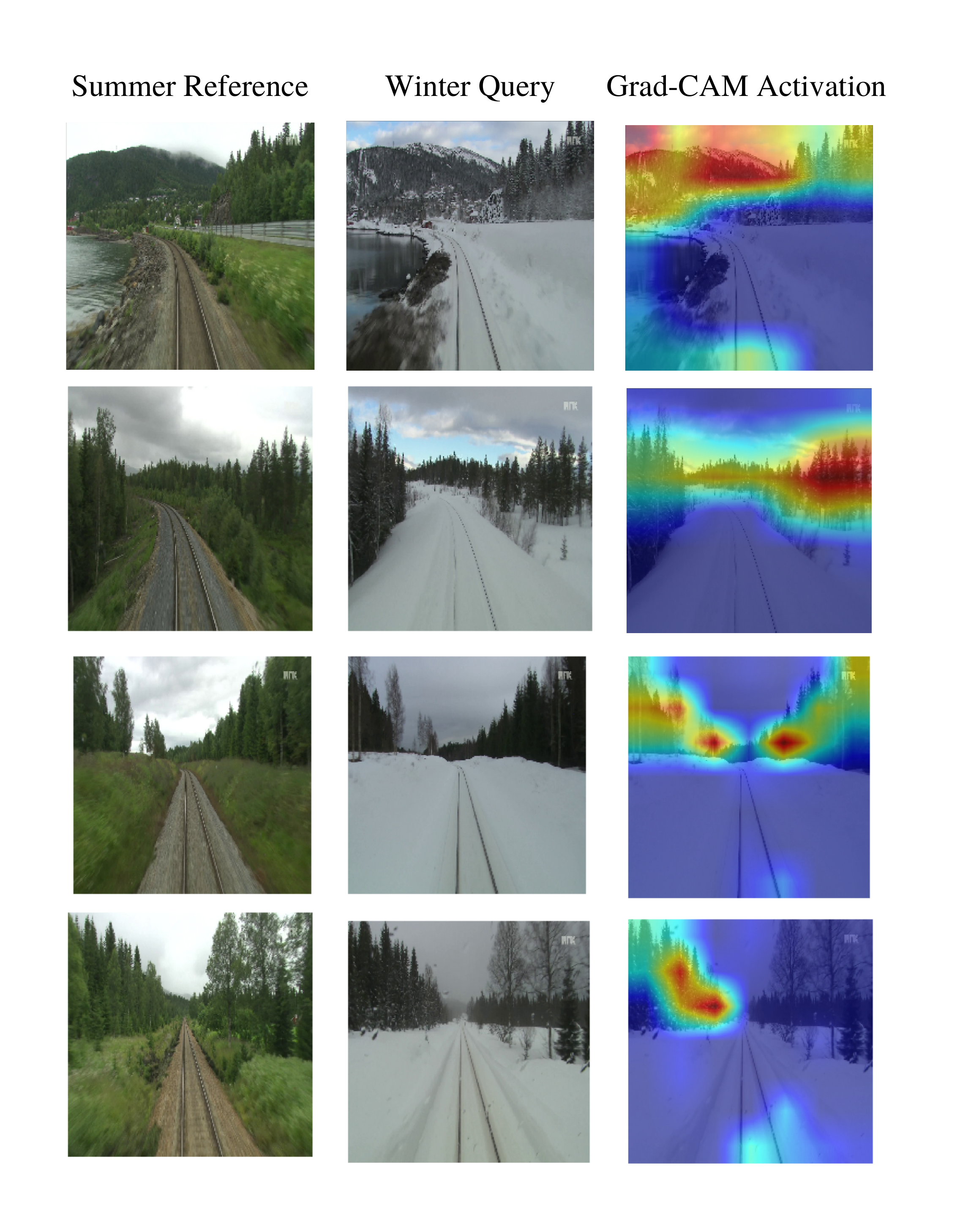}
    \caption{Visual Grad-CAM activation of input query winter image, along with retrieved summer image from the Nordland dataset.}
    \label{fig:nordland}
\end{figure}

Qualitative results are presented in Figure~\ref{fig:nordland} (Nordland dataset) and \ref{fig:Alderley} (Alderley). More qualitative results are included in supplementary materials. One can see examples of queries and best retrieved images, along with Grad-CAM~\cite{Selvaraju_2017_ICCV} activations. These visualizations demonstrate that CLASP-Net, even if trained without any labels, was able to learn features meaningful for outdoor localization tasks such as skylines for instance.

We focused our study on learning global visual representations that are robust to appearance changes and suitable for VPR. Our results demonstrate that it is possible to learn a model relying on constrastive self-supervision for robustness to appearance changes while being able to perceive the geometric structure of the input image by enforcing geometric prediction.



\begin{figure}
    \centering
    \includegraphics[trim=19mm 22mm 23mm 20mm, clip, width=\linewidth]{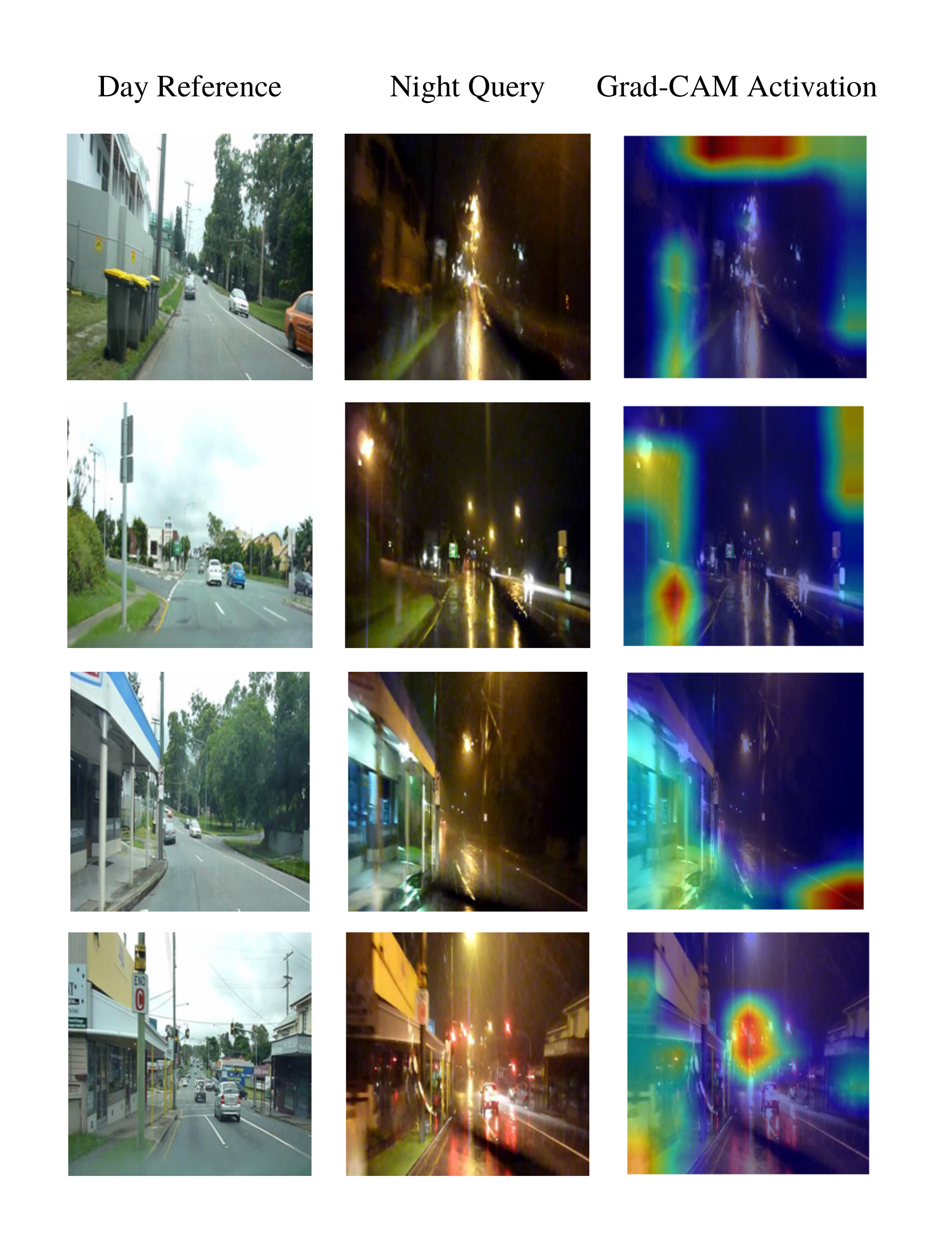}
    \caption{Visual Grad-CAM activation of input query night image, along with retrieved day image from the Alderley dataset.\vspace{-0.1cm}}
    \label{fig:Alderley}
\end{figure}


\subsection{Discussion on Potential Limitations}
\label{ssec:limits}


Global image descriptors typically offer greater robustness to environmental conditions at the expense of being less tolerant to viewpoint changes compared to local descriptors~\cite{lowry2015visual}. Our approach aims to further enhance the robustness to environmental conditions, allowing it to handle extreme scenarios as seen in the Nordland or Alderley datasets effectively. However, it is important to acknowledge that our method may encounter limitations when dealing with datasets that feature significant viewpoint variations between reference and query images for the same location, as our slightly weaker performance on the Oxford RobotCar dataset suggests.

\section{Conclusions}
\label{sec:conclusions}

In this paper, we presented CLASP-Net, a novel self-supervised approach designed for visual place recognition under challenging appearance variations. A significant advantage of our method is its independence from human supervision. CLASP-Net is trained to learn features that are both robust to appearance changes and sensitive to geometric nuances, serving as abstract place representations useful for visual place recognition tasks. Our extensive experimental evaluations substantiate the effectiveness and efficiency of the proposed approach. As a direction for future research, we aim to extend our model's capabilities by exploring sensitivity to 3D geometric transformations through view synthesis techniques.

{\small
\bibliographystyle{ieee_fullname}
\bibliography{main}
}

\end{document}



\title{Self-Supervised Learning for Place Representation Generalization across Appearance Changes\\-- Supplementary material --}

\author{Mohamed Adel Musallam\\
{\tt\small mohamed.ali@uni.lu}
\and Vincent Gaudilli\`ere\\
{\tt\small vincent.gaudilliere@uni.lu}
\and Djamila Aouada\\
{\tt\small djamila.aouada@uni.lu}
\and
SnT, University of Luxembourg
}

\maketitle
\ificcvfinal\thispagestyle{empty}\fi

In this supplementary material, we provide additional information, analyses and visualizations to support and enhance the understanding of our main paper. The main objective of our paper is to present a novel self-supervised learning framework that effectively generalizes place representations across various appearance changes in diverse environments. 

This supplementary material is organized as follows:

\textbf{Section 1}: We go through the technical details of our geometric augmentations for the self-supervised learning method, offering further explanation of the chosen type of transformation, the learning objectives and the optimization techniques employed. This aims at providing the readers with a thorough understanding of the inner working of our approach and its ability to generate robust place representations, as well as the reasoning behind our choices.

\textbf{Section 2}: We provide precisions regarding the reasoning behind appearance augmentations. 

\textbf{Section 3}: We share some qualitative visualizations of the learned place representations on the ``Oxford RobotCar Seasons v2" dataset, which offers valuable insights into how our framework captures meaningful features and adapts to appearance changes. These visualizations use Grad-CAM heatmaps and example images from different seasons allowing the readers to observe the effectiveness of our method in real-world scenarios.

\textbf{Section 4}: We present findings related to the hypothesis concerning the performance of CLASP-Net when the input query is subjected to rotation.

\section{Precisions on Geometric Augmentations}

\begin{figure}[t]
    \centering
    \includegraphics[width=\linewidth]{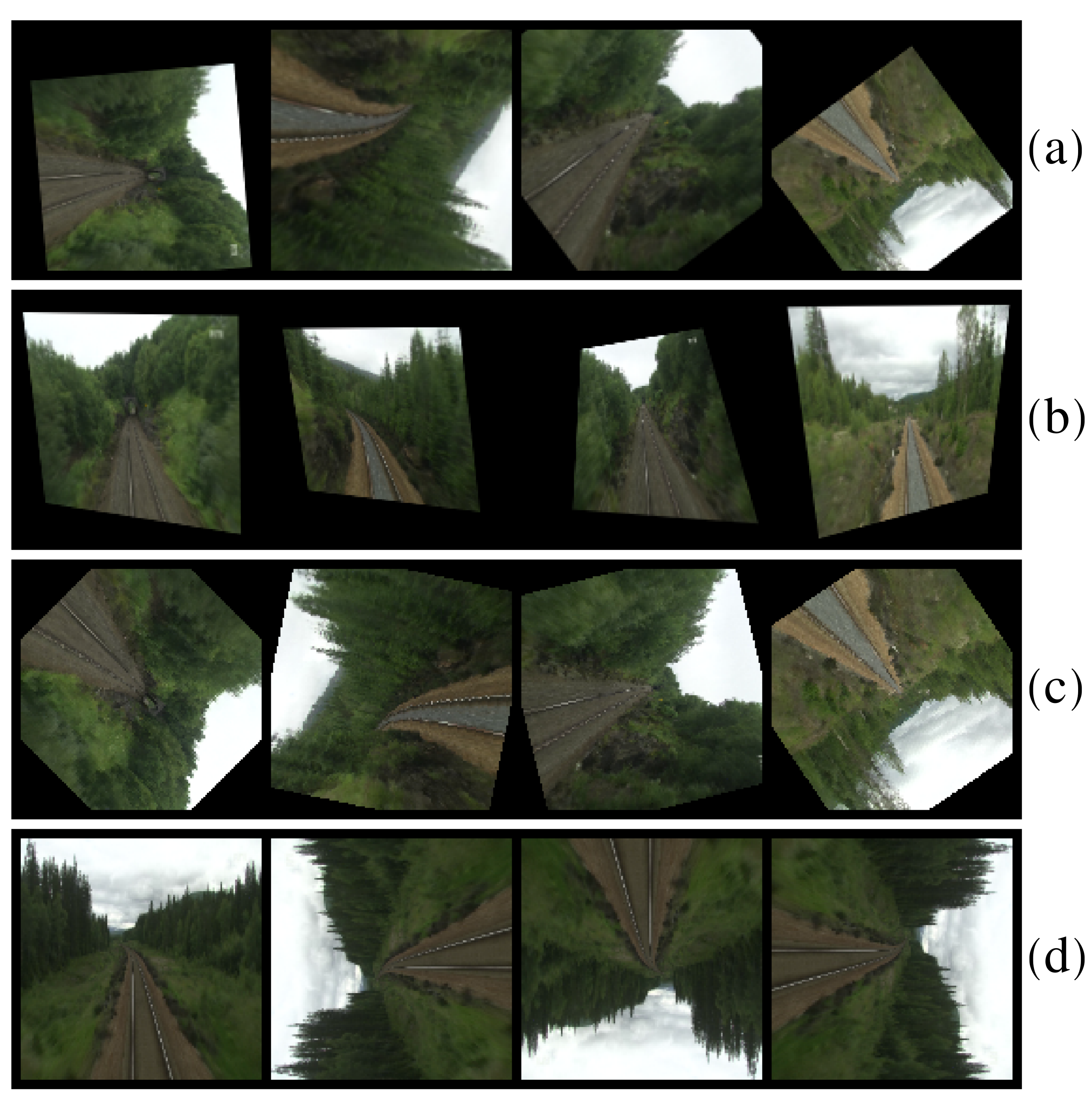}
    \caption{Examples of geometric augmentations: (a) affine transformations, (b) perspective transformations, (c) planar rotations by any angle, (d) planar rotations by 90\degree, 180\degree and 270\degree.}
    \label{fig:geometric_augmentation}
\end{figure}

\begin{figure}[t]
    \centering
    \includegraphics[trim=15cm 2cm 14cm 2.8cm, clip, width=.9\linewidth]{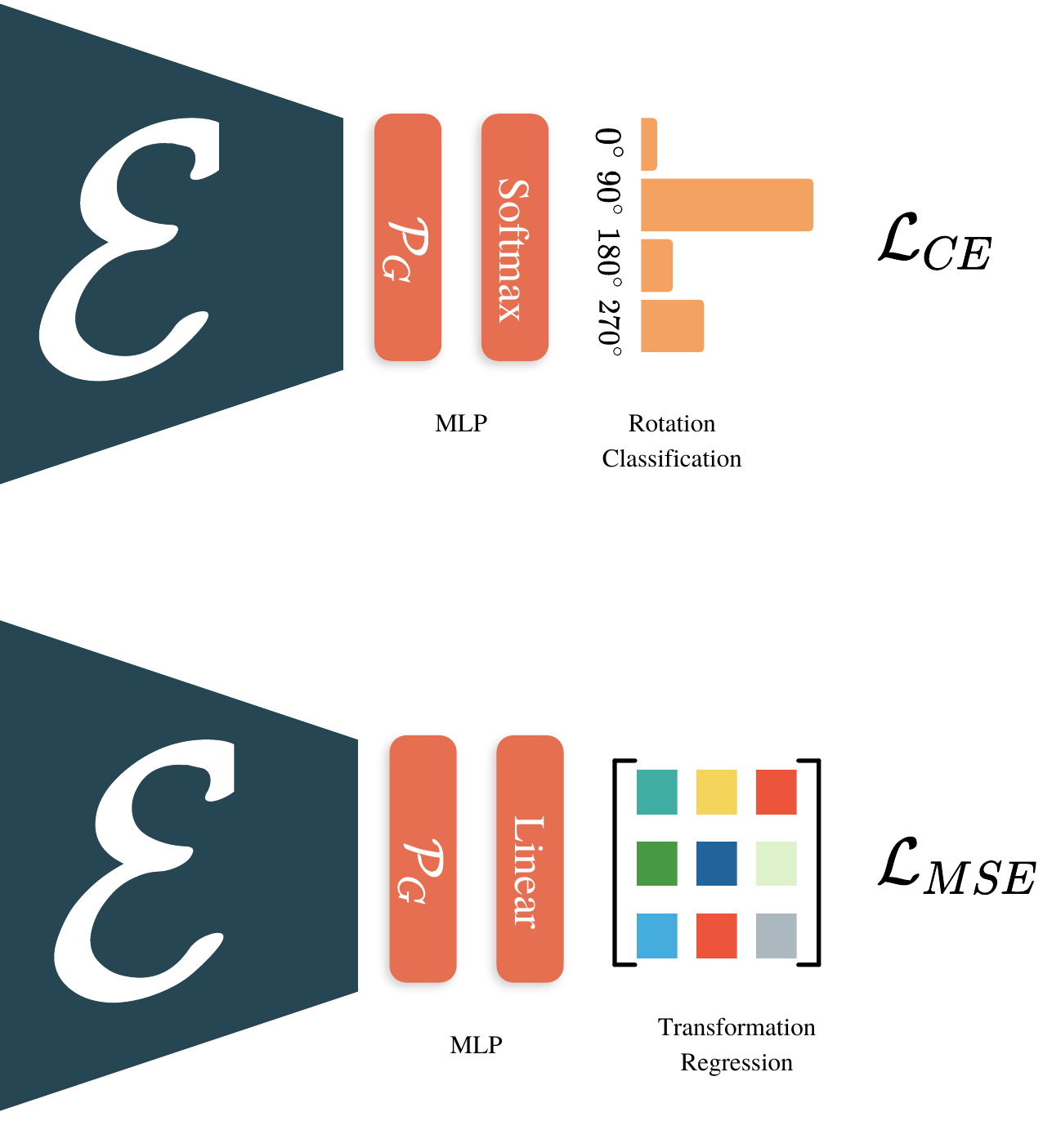}
    \caption{Illustrating the adaptation of the model to the different types of geometric augmentations.}
    \label{fig:2models}
\end{figure}

We tested different types of geometric augmentation (see Figure \ref{fig:geometric_augmentation}). In particular, random affine, random perspective, random rotation, and 90\degree-rotation have been used in our investigation. However, it has been observed that 90-degree rotation augmentation leads to better performance compared to the other types of augmentations.

There could be several reasons why the 90-degree rotation augmentation is more effective for visual place recognition.

Random affine and random perspective augmentations can distort the image's semantic content by changing the shape and size of objects, making it harder for the model to recognize the place, primarily since we rely on global features of the input image. 

On the other hand, random rotations and 90-degree rotations maintain the spatial structure of the image while providing variations in orientation. However, random continuous rotation may generate artifacts due to the boundary effect. Thus leveraging the group of 90-degree rotations is more helpful for the model to learn spatially meaningful representations, especially since there is no artifacts that the model can learn as shortcuts.

Furthermore, the effectiveness of different augmentations can depend on the dataset's specifics, as mentioned in~\cite[``Discussion" section]{dangovski2022equivariant}. For our case, natural visual navigation datasets contain a bias in the data towards up (sky) and down (ground), and breaking this natural setting without introducing other bias with the 90-degree rotation augmentation, then encouraging the network to be sensitive to this transformation, can help learning meaningful features.

In conclusion, the 90-degree rotation augmentation is more effective for visual place recognition than other tested geometric augmentations. This could be due to its ability to preserve the image's semantic content and spatial structure, provide variability in orientation, and disturb the natural uprigth setting of the places in the dataset.

The change in the model when learning to regress the parameters of the different types of geometric transformations is illustrated in Figure \ref{fig:2models}. In particular, the angle of the 90\degree-rotations is predicted as a classification problem and the training leverages the cross-entropy loss $\mathcal{L}_{CE}$, while the parameters of other transformations are regressed in their matrix form with a Mean Squared Error loss $\mathcal{L}_{MSE}$.

\begin{figure}[t]
    \centering
    \includegraphics[trim=19mm 22mm 23mm 20mm, clip, width=\linewidth]{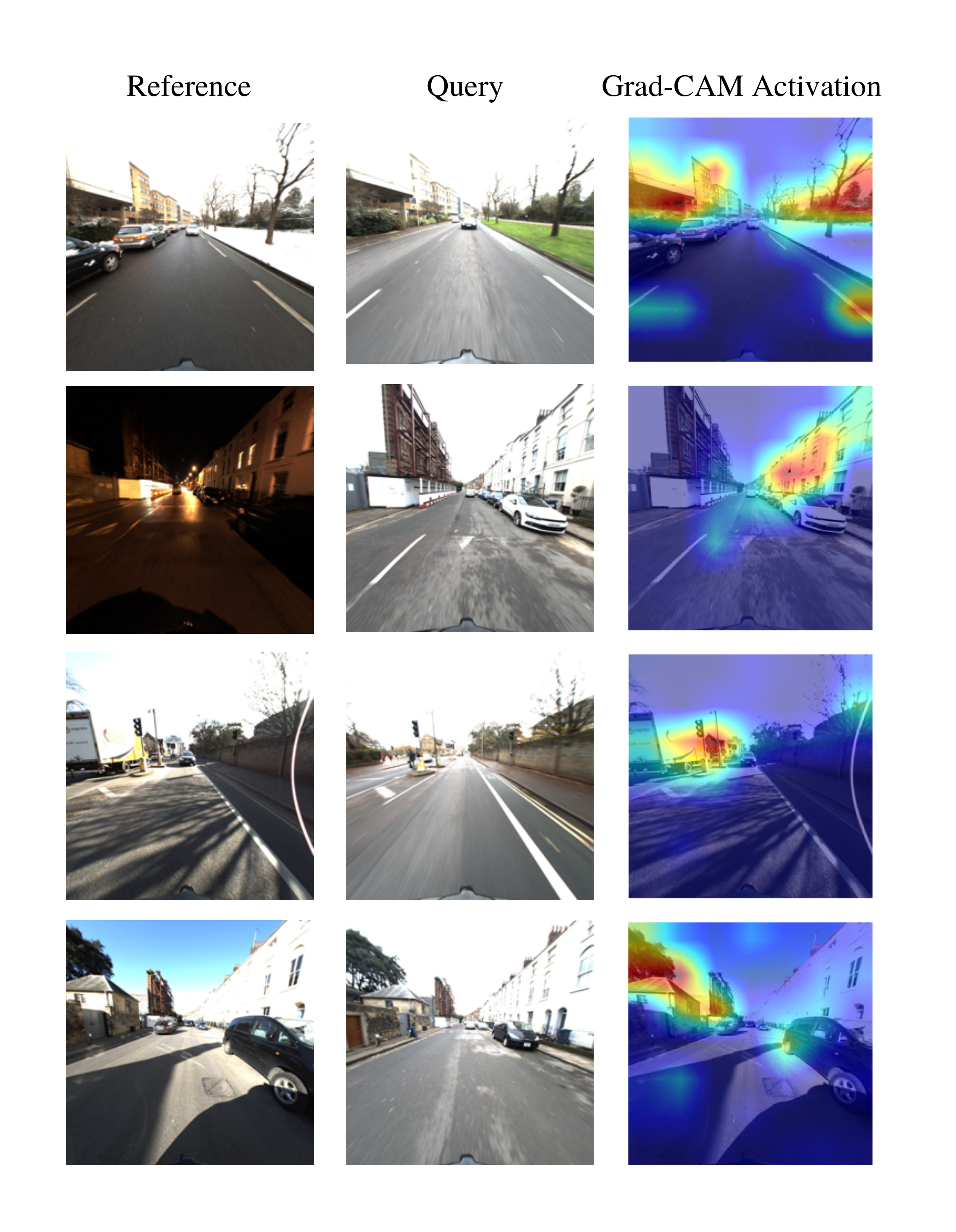}
    \caption{Visual grad-CAM activations of input query images, along with retrieved day images from the Oxford RobotCar Seasons v2 dataset.}
    \label{fig:robotcar}
\end{figure}

\section{Precisions on Appearance Augmentations}

In the context of visual navigation under appearance changes, pixel augmentation can be especially helpful in learning representations that are robust to changes in lighting, weather, and other factors that can affect the scene's appearance. In addition, by training our model using contrastive learning, the model can explicitly learn to map the same scene under different appearance variations to the same embedding, which can be crucial for successful navigation in real-world scenarios.

\section{Qualitative Results on Oxford RobotCar Seasons v2 dataset}

Figure \ref{fig:robotcar} shows qualitative results of CLASP-Net on the Oxford RobotCar Seasons v2 dataset, along with Grad-CAM activation maps.

\section{The Impact of Rotated Query Images}

While for visual place recognition, the orientation of the input image - particularly its natural up and down direction - is of critical significance, we intentionally break this inherent directionality to help the network learn distinctive locales and attributes of the visual environment. 

More precisely, in the rotation prediction task, to recognize a rotated version of a scene, the model needs to identify its different elements (\textit{e.g.} sky, ground, building, objects) and understand the spatial relationships between them~\cite{gidaris2018unsupervised}. Indeed, the GradCAM visualizations from our paper and supplementary material suggest that CLASP-Net learnt to detect geometric features such as skylines or buildings, which are helpful for both rotation prediction and outdoor place recognition (see \cite{SaurerBKLP16}).

However, we agree that such sensitivity to rotations may harm the place recognition performance under the same transformations of queries. Addressing such an hypothetical scenario, \textit{i.e.} recognizing the place depicted in a rotated query, therefore requires further adaptation of our original method. We have ran experiments to provide insights towards solving such a problem, whose results are presented below:
\begin{enumerate}[label=(\Alph*)]
    \item \textbf{Standard CLASP-Net on Unaltered Queries:}  When tested on unrotated queries, the baseline CLASP-Net yielded a performance score of \emph{$0.8025$}. This establishes the primary standard of comparison for the subsequent experiments.
    \item \textbf{Standard CLASP-Net on Rotated Queries:}  Upon challenging the CLASP-Net with rotated query images, with no retraining there was a marked decrease in performance, registering a score of \emph{$0.3729$} This significant drop underscores the sensitivity of the model to rotations in its standard configuration.
     \item \textbf{CLASP-Net with Rotation Augmentation in the Invariance Module on Unaltered Queries:} By integrating rotation augmentation in the invariance module, the performance on unrotated images slightly diminished to \textbf{0.7624}. This suggests that while the model becomes better suited to handle rotations, there might be a slight compromise in its efficacy on standard queries.

     \item  \textbf{CLASP-Net with Rotation Augmentation in the Invariance Module on Rotated Queries:} Significantly, this configuration exhibited a performance score of  \textbf{0.7744} on rotated queries. This is a stark contrast to the performance of the standard CLASP-Net on rotated queries, highlighting the efficacy of the rotation augmentation in bolstering the model's resilience against rotations.

     \item  \textbf{CLASP-Net Excluding the Rotation Prediction Module with Rotation Augmentation in the Invariance Module on Unaltered Queries:}  Omitting the equivariance module while retaining the rotation augmentation in the invariance module resulted in a performance of \textbf{0.7872} on standard queries.

     \item \textbf{CLASP-Net Excluding the Rotation Prediction Module with Rotation Augmentation in the Invariance Module on Rotated Queries:}  This configuration produced a score of \textbf{0.712} on rotated queries. While this is lower than the performance achieved with the rotation-augmented CLASP-Net on rotated queries, it is considerably better than the standard CLASP-Net's performance on similar inputs.

\end{enumerate}

\begin{figure}
    \centering
    \includegraphics[width=\linewidth]{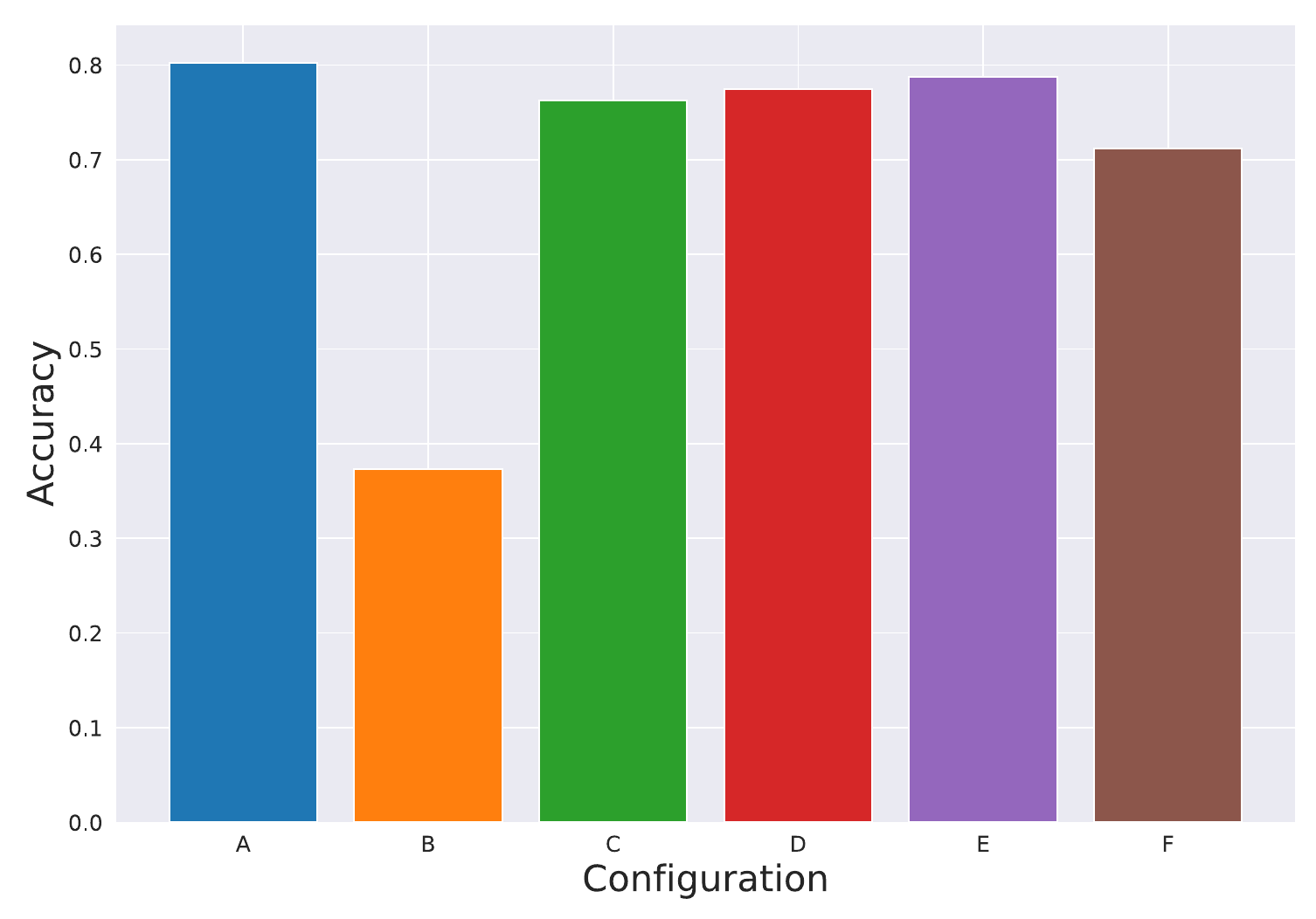}
    \caption{Testing the model with a rotated query in a different configuration on the Nordland Summer/Winter $R @ 10$ .}
    \label{fig:enter-label}
\end{figure}

Our training strategy encourages information about rotations to be retained in the image representation rather than guaranteeing strict equivariance. However, in practice, we observe that the average cosine similarity between representations of rotated views (referred to as \textit{equivariant measure} in \cite{dangovski2022equivariant}) tends to 0 (\textit{i.e.}, 90$\degree$ angle) when the dedicated module is added (see Table \ref{tab:ablation}). Moreover, the choice of this particular group of geometric transformations is the outcome of experimentation. In particular, it shows that the best performance is achieved with the cyclic group of 90$\degree$ rotations, compared to the groups of 2D affine transformations, 2D projective transformations, and 2D rotations.

 \begin{table}[t]
\centering
\small
\begin{tabular}{l|c|c}
\hline
Module                      & R@10 & Equiv. meas.\cite{dangovski2022equivariant_short}\\ \hline
Pre-trained encoder             & 28.2\% & 0.339         \\
Appearance Module & 75.8\% & 0.189         \\
Geometry Module & 52.3\% & \textbf{0.093}         \\
Combined modules (ACM-Net)          & \textbf{80.2\%} & 0.139         \\ \hline
\end{tabular}
\caption{CLASP-Net analysis on Nordland summer/winter.}
\label{tab:ablation}
\end{table}

In summary, integrating rotation augmentation within the invariance module substantially improves CLASP-Net's capability to handle rotated queries. However, the omission of the equivariance module has nuanced effects; These insights are instrumental for refining architectures based on the downstream task and the assumption about the input images and enhancing the robustness of the models in the face of varied input conditions.

{\small
\bibliographystyle{ieee_fullname}
\bibliography{egbib}
}